\pdfoutput=1
\documentclass[11pt]{article}
\usepackage[letterpaper,margin=1in]{geometry}

\usepackage{lmodern}
\usepackage[T1]{fontenc}
\usepackage{amsmath}
\usepackage{booktabs}
\usepackage{url}
\usepackage{array}
\usepackage{graphicx}
\usepackage[tableposition=top]{caption}
\usepackage[numbers,sort&compress]{natbib}
\usepackage[hidelinks]{hyperref}
\usepackage{orcidlink}
\hypersetup{
  pdftitle={Environmental sensor readings in two crop disease image datasets identify the session in which each image was taken},
  pdfauthor={Sungwoo Kang}}

\newcommand{\numImages}{274{,}000}
\newcommand{\numCrops}{8}
\newcommand{\numClasses}{17}
\newcommand{\numSingleCropClasses}{12}
\newcommand{\numFarms}{24}
\newcommand{\numPanelTuples}{13{,}150}
\newcommand{\numTuplesOneSession}{13{,}144}
\newcommand{\photoWindow}{August--November 2023}
\newcommand{\pctPhotoDayDiffers}{9.3}

\newcommand{\rsqFarmLo}{0.83}
\newcommand{\rsqFarmHi}{0.91}
\newcommand{\pctPanelTupleShared}{99.5}
\newcommand{\pctPhImpossible}{15.9}
\newcommand{\pctNpkZero}{26.8}

\newcommand{\numSeeds}{7}
\newcommand{\timestampCrops}{7}
\newcommand{\comparableBand}{0.02}
\newcommand{\basePooledPlain}{0.359}
\newcommand{\basePooledBA}{0.255}
\newcommand{\panelPooledPlain}{0.643}
\newcommand{\timestampPooledPlain}{0.677}
\newcommand{\panelPooledBA}{0.596}
\newcommand{\timestampPooledBA}{0.637}
\newcommand{\basePooledPlainCI}{[0.359,\,0.360]}
\newcommand{\basePooledBACI}{[0.255,\,0.255]}
\newcommand{\panelPooledPlainCI}{[0.642,\,0.644]}
\newcommand{\timestampPooledPlainCI}{[0.676,\,0.678]}
\newcommand{\panelPooledBACI}{[0.595,\,0.596]}
\newcommand{\timestampPooledBACI}{[0.636,\,0.638]}
\newcommand{\timestampGeqCrops}{7}
\newcommand{\panelDecisiveCrops}{0}
\newcommand{\figWinPlain}{+0.154}
\newcommand{\figWinCI}{[+0.153,\,+0.155]}
\newcommand{\mangoWinPlain}{+0.060}
\newcommand{\mangoWinCI}{[+0.059,\,+0.062]}
\newcommand{\oliveLeanPlain}{-0.020}
\newcommand{\oliveLeanCI}{[-0.022,\,-0.018]}
\newcommand{\oliveLeanPlainFine}{-0.0196}
\newcommand{\panelAdvantagePooled}{0.251}
\newcommand{\panelAdvantageWithinSession}{0.045}
\newcommand{\residualLo}{0.027}
\newcommand{\residualHi}{0.071}
\newcommand{\figWinBA}{+0.185}
\newcommand{\figWinBACI}{[+0.183,\,+0.187]}
\newcommand{\mangoWinBA}{+0.070}
\newcommand{\mangoWinBACI}{[+0.069,\,+0.072]}
\newcommand{\oliveLeanBA}{-0.018}
\newcommand{\oliveLeanBACI}{[-0.024,\,-0.013]}

\newcommand{\fusionSeeds}{5}
\newcommand{\subsampleSize}{40{,}000}
\newcommand{\lowBudgetEpochs}{40}
\newcommand{\matchedEpochs}{300}
\newcommand{\matchedBatch}{8}
\newcommand{\imageOnlyFone}{0.823}
\newcommand{\imageOnlyFoneMatched}{0.809}
\newcommand{\imageOnlyFoneDate}{0.762}
\newcommand{\imageOnlyFoneDateMatched}{0.759}
\newcommand{\iidDeltaLowBudget}{+0.0013}
\newcommand{\iidDeltaLowBudgetCI}{[-0.0058,\,+0.0084]}
\newcommand{\dateDeltaLowBudget}{+0.0082}
\newcommand{\dateDeltaLowBudgetCI}{[-0.0091,\,+0.0255]}
\newcommand{\imageOnlyFoneFarm}{0.323}
\newcommand{\farmDeltaLowBudget}{+0.0465}
\newcommand{\farmDeltaLowBudgetCI}{[-0.0053,\,+0.0983]}
\newcommand{\iidDelta}{+0.0084}
\newcommand{\iidDeltaCI}{[+0.0028,\,+0.0140]}
\newcommand{\imageOnlyBudgetChange}{-0.0141}
\newcommand{\fusionBudgetChange}{-0.0071}
\newcommand{\dateDelta}{+0.0037}
\newcommand{\dateDeltaCI}{[-0.0041,\,+0.0115]}
\newcommand{\dateReduction}{+0.0046}
\newcommand{\dateReductionCI}{[-0.0010,\,+0.0103]}
\newcommand{\iidDeltaLowBudgetBA}{+0.0129}
\newcommand{\iidDeltaLowBudgetBACI}{[+0.0110,\,+0.0149]}
\newcommand{\dateDeltaLowBudgetBA}{+0.0171}
\newcommand{\dateDeltaLowBudgetBACI}{[+0.0038,\,+0.0303]}
\newcommand{\farmDeltaLowBudgetBA}{+0.0493}
\newcommand{\farmDeltaLowBudgetBACI}{[-0.0010,\,+0.0996]}
\newcommand{\iidDeltaBA}{+0.0073}
\newcommand{\iidDeltaBACI}{[+0.0002,\,+0.0145]}
\newcommand{\dateDeltaBA}{+0.0121}
\newcommand{\dateDeltaBACI}{[+0.0029,\,+0.0212]}
\newcommand{\dateReductionBA}{-0.0048}
\newcommand{\dateReductionBACI}{[-0.0153,\,+0.0058]}
\newcommand{\iidPanelLeak}{89.8}
\newcommand{\datePanelLeak}{7.3}
\newcommand{\farmCellPos}{35.1}
\newcommand{\farmCellPosMean}{72.7}
\newcommand{\positivityFloor}{80}

\newcommand{\gateAxes}{4}
\newcommand{\gateFolds}{5}
\newcommand{\gateRandomLeak}{99.3}
\newcommand{\gateDateLeak}{7.2}
\newcommand{\gateSessionLeak}{7.7}
\newcommand{\gateFarmCellPos}{72.6}

\newcommand{\cddImages}{5{,}767}
\newcommand{\cddSeries}{780}
\newcommand{\cddSharedPct}{96.4}
\newcommand{\cddSingleDatePct}{98.6}
\newcommand{\cddSingleCodePct}{98.4}
\newcommand{\cddDates}{64}
\newcommand{\cddSeriesPerDate}{12.2}
\newcommand{\cddTransferableImages}{35}

\newcommand{\cddLeak}{91.9}
\newcommand{\cddLeakMapped}{91.0}
\newcommand{\cddLeakTrainval}{96.4}
\newcommand{\cddGroupsShared}{406}
\newcommand{\cddGroupsTotal}{485}
\newcommand{\cddSecondSplitLeak}{97.1}
\newcommand{\cddOtherLeakTrainval}{99.1}
\newcommand{\cddDirectTest}{31.5}
\newcommand{\cddDirectTrainval}{50.2}

\newcommand{\cddSeeds}{7}
\newcommand{\cddEnvRandom}{0.885}
\newcommand{\cddEnvRandomCI}{[0.873,\,0.896]}
\newcommand{\cddEnvGroup}{0.653}
\newcommand{\cddCollapseBand}{0.03}
\newcommand{\cddEnvGroupCI}{[0.587,\,0.720]}
\newcommand{\cddTimestampRandom}{0.899}
\newcommand{\cddTimestampRandomCI}{[0.886,\,0.912]}
\newcommand{\cddTimestampGroup}{0.595}
\newcommand{\cddTimestampGroupCI}{[0.556,\,0.633]}
\newcommand{\cddBaseRandom}{0.014}
\newcommand{\cddBaseRandomCI}{[0.014,\,0.014]}
\newcommand{\cddBaseGroup}{0.015}
\newcommand{\cddBaseGroupCI}{[0.015,\,0.016]}
\newcommand{\cddTimestampEnvRandom}{0.914}
\newcommand{\cddTimestampEnvRandomCI}{[0.910,\,0.917]}
\newcommand{\cddTimestampEnvGroup}{0.673}
\newcommand{\cddTimestampEnvGroupCI}{[0.591,\,0.754]}
\newcommand{\cddTimestampEnvOverTimestamp}{+0.015}
\newcommand{\cddTimestampEnvOverTimestampCI}{[+0.003,\,+0.026]}
\newcommand{\cddTimestampEnvOverEnv}{+0.029}
\newcommand{\cddTimestampEnvOverEnvCI}{[+0.018,\,+0.040]}
\newcommand{\cddTimestampOverEnvRandom}{+0.014}
\newcommand{\cddTimestampOverEnvRandomCI}{[-0.003,\,+0.031]}
\newcommand{\cddTimestampOverEnvGroup}{-0.059}
\newcommand{\cddTimestampOverEnvGroupCI}{[-0.119,\,+0.001]}
\newcommand{\cddLeakRandom}{95.9}
\newcommand{\cddLeakGroup}{0.0}
\newcommand{\cddBaseRandomBA}{0.040}
\newcommand{\cddBaseRandomBACI}{[0.040,\,0.040]}
\newcommand{\cddEnvRandomBA}{0.884}
\newcommand{\cddEnvRandomBACI}{[0.872,\,0.896]}
\newcommand{\cddTimestampRandomBA}{0.893}
\newcommand{\cddTimestampRandomBACI}{[0.879,\,0.907]}
\newcommand{\cddTimestampEnvRandomBA}{0.910}
\newcommand{\cddTimestampEnvRandomBACI}{[0.907,\,0.914]}
\newcommand{\cddTimestampOverEnvRandomBA}{+0.009}
\newcommand{\cddTimestampOverEnvRandomBACI}{[-0.009,\,+0.027]}
\newcommand{\cddBaseGroupBA}{0.045}
\newcommand{\cddBaseGroupBACI}{[0.043,\,0.046]}
\newcommand{\cddEnvGroupBA}{0.695}
\newcommand{\cddEnvGroupBACI}{[0.638,\,0.752]}
\newcommand{\cddTimestampGroupBA}{0.637}
\newcommand{\cddTimestampGroupBACI}{[0.591,\,0.682]}
\newcommand{\cddTimestampEnvGroupBA}{0.720}
\newcommand{\cddTimestampEnvGroupBACI}{[0.651,\,0.788]}
\newcommand{\cddTimestampOverEnvGroupBA}{-0.058}
\newcommand{\cddTimestampOverEnvGroupBACI}{[-0.102,\,-0.014]}
\newcommand{\cddTheirSeeds}{7}
\newcommand{\cddTheirBase}{0.0135}
\newcommand{\cddTheirBaseCI}{[0.0135,\,0.0135]}
\newcommand{\cddTheirEnv}{0.8450}
\newcommand{\cddTheirEnvCI}{[0.8417,\,0.8484]}
\newcommand{\cddTheirTimestamp}{0.9095}
\newcommand{\cddTheirTimestampCI}{[0.9035,\,0.9156]}
\newcommand{\cddTheirTimestampEnv}{0.8851}
\newcommand{\cddTheirTimestampEnvCI}{[0.8821,\,0.8882]}
\newcommand{\cddTheirTimestampOverTimestampEnv}{+0.0244}
\newcommand{\cddTheirTimestampOverTimestampEnvCI}{[+0.0187,\,+0.0301]}
\newcommand{\cddOtherBase}{0.0135}
\newcommand{\cddOtherBaseCI}{[0.0135,\,0.0135]}
\newcommand{\cddOtherEnv}{0.8910}
\newcommand{\cddOtherEnvCI}{[0.8889,\,0.8931]}
\newcommand{\cddOtherTimestamp}{0.9450}
\newcommand{\cddOtherTimestampCI}{[0.9418,\,0.9481]}
\newcommand{\cddOtherTimestampEnv}{0.9421}
\newcommand{\cddOtherTimestampEnvCI}{[0.9379,\,0.9464]}
\newcommand{\cddOtherTimestampOverTimestampEnv}{+0.0028}
\newcommand{\cddOtherTimestampOverTimestampEnvCI}{[+0.0002,\,+0.0054]}
\newcommand{\cddTheirBaseBA}{0.0400}
\newcommand{\cddTheirBaseBACI}{[0.0400,\,0.0400]}
\newcommand{\cddTheirEnvBA}{0.8452}
\newcommand{\cddTheirEnvBACI}{[0.8430,\,0.8474]}
\newcommand{\cddTheirTimestampBA}{0.8962}
\newcommand{\cddTheirTimestampBACI}{[0.8894,\,0.9031]}
\newcommand{\cddTheirTimestampEnvBA}{0.8861}
\newcommand{\cddTheirTimestampEnvBACI}{[0.8824,\,0.8897]}
\newcommand{\cddTheirTimestampOverTimestampEnvBA}{+0.0102}
\newcommand{\cddTheirTimestampOverTimestampEnvBACI}{[+0.0033,\,+0.0171]}
\newcommand{\cddOtherBaseBA}{0.0400}
\newcommand{\cddOtherBaseBACI}{[0.0400,\,0.0400]}
\newcommand{\cddOtherEnvBA}{0.8925}
\newcommand{\cddOtherEnvBACI}{[0.8906,\,0.8944]}
\newcommand{\cddOtherTimestampBA}{0.9442}
\newcommand{\cddOtherTimestampBACI}{[0.9411,\,0.9474]}
\newcommand{\cddOtherTimestampEnvBA}{0.9411}
\newcommand{\cddOtherTimestampEnvBACI}{[0.9381,\,0.9440]}
\newcommand{\cddOtherTimestampOverTimestampEnvBA}{+0.0031}
\newcommand{\cddOtherTimestampOverTimestampEnvBACI}{[+0.0021,\,+0.0042]}

\newcommand{\cddCrops}{6}
\newcommand{\cddClasses}{25}
\newcommand{\cddSingleDateClasses}{4}

\newcommand{\forestTrees}{60}
\newcommand{\timestampMinRows}{1{,}000}
\newcommand{\timestampSampleRows}{20{,}000}
\newcommand{\timestampFolds}{3}
\newcommand{\sessionMinRows}{300}
\newcommand{\sessionMinClassRows}{3}
\newcommand{\cddFolds}{7}
\newcommand{\fusionFolds}{5}
\newcommand{\imagePixels}{256}
\newcommand{\lstmHidden}{200}
\newcommand{\learningRate}{0.001}
\newcommand{\restartEpochs}{25}
\newcommand{\lowBudgetBatch}{64}
\newcommand{\earlyStopPatience}{5}

\newcommand{\pythonVersion}{3.12}
\newcommand{\pandasVersion}{3.0.3}
\newcommand{\sklearnVersion}{1.9.0}
\newcommand{\torchVersion}{2.13.0}
\newcommand{\timmVersion}{1.0.28}
\newcommand{\albumentationsVersion}{2.0.8}

\newcommand{\paperIncrement}{2.58}
\newcommand{\paperImageFone}{0.8779}
\newcommand{\paperMultiFone}{0.9037}

\begin{document}

\title{Environmental sensor readings in two crop disease image datasets identify
the session in which each image was taken}

\author{Sungwoo Kang\texorpdfstring{\,\orcidlink{0009-0004-0037-7593}}{}\\[2pt]
Department of Electrical and Computer Engineering, Korea University\\
Seoul, Republic of Korea\\
\texttt{krml919@korea.ac.kr}}
\date{}

\maketitle

\begin{abstract}
\textbf{Introduction:} Crop disease studies that fuse images with environmental
sensor readings, such as soil chemistry, report higher accuracy than image-only
models. In a paired image--sensor dataset, however, one reading can be attached to
many images of a session, which is one farm or greenhouse on one day. If disease
classes differ between sessions, the reading can predict the class by identifying the
session rather than by measuring the plant's environment.
\textbf{Methods:} We tested whether the sensor readings of two Korean datasets
identify the session, using a diagnostic that reads no image. The datasets are the Crop Disease Diagnosis (CDD) challenge dataset and a
subtropical and tropical crop pest and disease dataset from the government portal
AI Hub. The diagnostic
counts how many images share each sensor reading and compares a classifier given the sensor readings with one given
only the date and time that the dataset records. We
also re-analyzed a published CDD fusion study and trained its fusion model on the
AI Hub dataset.
\textbf{Results:} In both datasets, nearly every image shared its sensor reading
with other images. On the AI Hub dataset, the date and time alone were comparable
to or better than the dataset's soil-chemistry readings at predicting the disease
class in all \timestampCrops{} crops that we compared. For \cddLeak{}\% of the CDD
study's test images, a training image carried identical sensor readings. On the
study's split, the date and time alone matched the study's reported macro-F1 for
its fusion model. Trained for the study's \matchedEpochs{} epochs, the
fusion model raised macro-F1 slightly relative to an image-only model on a random
split of the AI Hub dataset, but whether the gain exceeds zero on
held-out dates remained unresolved.
\textbf{Discussion:} A fusion gain measured on a split that shares sessions
between training and test data cannot be separated from session identity. Fusion
studies should therefore report the gain on a split that holds out whole farms or
sessions, and should set a sensor-only score beside those of two predictors that
read no sensor: one given only the date and time and one that always outputs the
most frequent class.

\medskip\noindent\textbf{Keywords:} plant disease detection, multimodal fusion, environmental sensors, data leakage, shortcut learning, cross-validation, benchmark datasets
\end{abstract}

\section{Introduction}
\label{sec:intro}

Plant pathogens and pests reduce the yields of the major food crops worldwide
\citep{savary2019burden}. To detect plant diseases, deep convolutional networks have been trained to
recognize them in leaf images \citep{mohanty2016deep}. Recent studies add environmental sensor
readings, such as air temperature, humidity, and soil chemistry recorded by
Internet of Things devices, to the image in one multimodal model and report that
the fusion model detects diseases and pests more accurately than an image-only
model \citep{xu2025wheat,jamun2025iot,partitioned2026}. Datasets released for
research supply paired images and sensor readings for training fusion models. A dataset of
subtropical and tropical crop pests and diseases on AI Hub, the Korean
government's data portal, attaches a panel of soil and ambient sensor variables to
each image and lists predicting disease from that panel among its intended uses
\citep{aihub2023subtropical}. The Crop Disease Diagnosis (CDD) challenge dataset
pairs each labeled image with an environment series, a time series of greenhouse
sensor readings \citep{dacon2021cropdisease}. A study on CDD reports that adding
the environment series to a ResNet-50 image classifier raises macro-F1 by
\paperIncrement{} percentage points \citep{appliedsci2024dacon}. Whether such
gains come from the plant's environment depends on what the paired sensor
readings measure.

\begin{figure}[tbp]
\centering
\includegraphics[width=\textwidth]{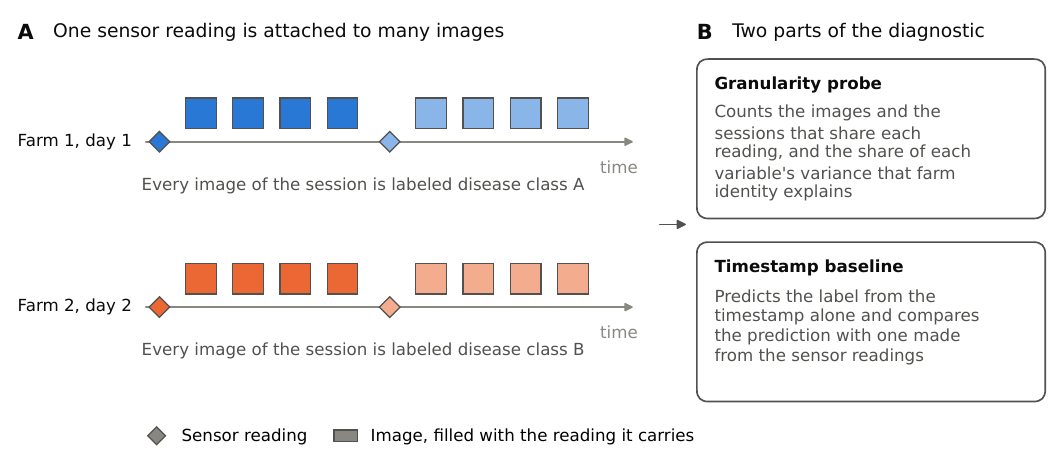}
\caption{How one sensor reading is attached to many images, and the two parts of
the diagnostic. \textbf{(A)} A schematic of two sessions, each one farm on one day.
Each sensor reading (diamond) is attached to every image (square) taken after it
until the next reading, so the images of a session share far fewer readings than there are images. In the schematic, every image
of a session carries that session's disease class. \textbf{(B)} The two parts of
the diagnostic. The granularity probe counts the images and the sessions that
share each reading and measures the share of each variable's variance that farm
identity explains. The timestamp baseline compares a classifier given only the
timestamp with one given the sensor readings. Neither part reads an image.}
\label{fig:concept}
\end{figure}

A sensor at a farm records a reading at intervals, so a dataset can attach one
reading to every image taken at that farm before the next reading
(Figure~\ref{fig:concept}A). A reading attached in this way identifies the
farm and the day of acquisition. We call one farm or greenhouse on one day a
session. If disease classes differ between the sessions of a
dataset, a model can predict the class by identifying the session from the
readings, without any agronomic signal in them. The readings are then a shortcut,
which is a feature that predicts the label in the training data but not under new
conditions \citep{geirhos2020shortcut,koh2021wilds}. A random split, however,
tests a model under the conditions of its training data: it places images of one
session in both the training and the test data, so a model can learn in training which class goes
with the readings of each session and recognize the same readings in the test data. A fusion gain measured on a random split therefore cannot separate
an agronomic contribution of the readings from session identity, which makes the session
a confound of the gain.

Session-disjoint splits, which keep all images of a session on one side and so
would separate the agronomic contribution from session identity, already exist for image
datasets. Image datasets need such splits because the images of one session share
the conditions under which they were taken, so images identify their session
as well. A classifier's accuracy on the sessions of its training data therefore
need not hold on other sessions: classifiers trained on images taken under controlled
conditions lose accuracy on images taken under other conditions
\citep{mohanty2016deep,wu2023lab2field}. The Global Wheat Head Detection benchmark,
for example, supplies session-disjoint splits that measure that loss. It divides its images into sub-datasets, each of which holds
the images of one experimental unit from one acquisition session \citep{gwhd2021}. A
sub-dataset therefore corresponds to one of our sessions, with its experimental
unit as the farm or greenhouse and its acquisition session as the day. The benchmark assigns
each sub-dataset to only one of its training, validation, and test sets
\citep{gwhd2021}. Its splits therefore measure
accuracy on sessions that the training data do not include. More generally, for data with spatial, temporal, or
hierarchical structure, \citet{roberts2017crossvalidation} recommend
cross-validation that holds out whole blocks of related observations. Session
identity is less apparent in sensor readings than in images, however, because a
sensor reading is a physical measurement and so appears to carry agronomic
information.

The fusion studies that we cite
\citep{xu2025wheat,jamun2025iot,agrovision2025,agrifusion2026,cornsentry2026,appliedsci2024dacon,berrocal2026,partitioned2026}
apply neither of two checks that would show whether their sensor readings
identify the session. A timestamp, which
is the date and time that a dataset records for each image or for its sensor
reading, identifies the day directly, yet none of the studies compares how well its
sensor readings and a timestamp predict the label. None of the studies reports an
evaluation on a farm-disjoint or session-disjoint split, which keeps all images of
a farm or of a session in either the training or the test data. One of the studies,
however, keeps all image and sensor sequences of each plant or each infection
event (one occurrence of disease) in only one of its training,
validation, and test sets \citep{partitioned2026}. That study
states a temporal separation between its training and test data, but its text
does not establish that no session falls on both sides.

We supply a diagnostic that tests whether a dataset's sensor readings identify the
session. The diagnostic reads a dataset's sensor
readings and the metadata that the dataset records for each image, but no image.
Testing a fusion model on held-out farms or sessions, by contrast, needs the
trained model and
needs every class to occur on several farms or sessions, so that each class
appears in both the training and the test data. The diagnostic needs neither, so
it can run before any fusion model is trained. The
diagnostic has two parts (Figure~\ref{fig:concept}B). A granularity probe
tests whether the sensor readings describe single images or groups of images: it
measures how many images share each sensor reading, how many sessions each
reading occurs in, and the share of each sensor
variable's variance that farm identity explains. A timestamp baseline predicts the
label from a timestamp alone and compares that prediction with one made from the sensor readings, because a
reading that predicts the label no better than the timestamp may only identify the
session. The timestamp baseline adapts partial-input baselines,
which score a classifier that sees only part of its input and so measure how well
that part alone predicts the label \citep{feng2019misleading}. Such baselines
have been applied to natural language inference, in which a model decides whether
the second sentence of a pair follows from the first. On the benchmarks of that
task, classifiers that saw only the second sentence still predicted the label, so
the second sentences carried cues to the label
\citep{poliak2018hypothesis,gururangan2018artifacts}. Unlike a partial-input
baseline, which keeps part of the model's input, the timestamp baseline replaces
the sensor readings with the timestamp, so the baseline measures how well the
timestamp alone predicts the label.

We apply the diagnostic to the AI Hub and CDD datasets and re-analyze the CDD
study on its own split, because the CDD study is the only fusion study that we
cite whose dataset and split were both released. The CDD study also supplies the
fusion model that we train on the AI Hub dataset to test whether the model's gain
exceeds zero when whole photo dates are held out. Holding out photo dates keeps
the session of most test images out of training, because most images of the AI
Hub dataset were photographed on the day on which their panel was measured. The
test runs on the AI Hub dataset because, unlike CDD, it can be
split by photo date with every class in both the training and the test data. This
paper makes five contributions.
\begin{itemize}
  \item In both datasets, one sensor reading is shared by many images: the AI Hub
  dataset carries only \numPanelTuples{} distinct sets of panel values across a
  metadata spreadsheet of \numImages{} rows, one row per image, and CDD only
  \cddSeries{} distinct environment series for its \cddImages{} images.
  \item On the AI Hub dataset, a classifier given only the timestamp is
  comparable to or better than one given the dataset's soil-chemistry
  readings in
  \timestampGeqCrops{} of the \timestampCrops{} crops compared.
  \item On the CDD study's own split, where \cddLeak{}\% of the test images share
  their environment series with training images, a classifier given only the
  timestamp reaches the macro-F1 that the study reports for its fusion model.
  \item On CDD, a classifier given only the environment series falls in macro-F1
  from \cddEnvRandom{} on a random split to \cddEnvGroup{} on a split that keeps
  all images of each series in either the training or the test data, so much of its random-split score comes from
  recognizing series seen in training. Under either split, we detect no
  difference in macro-F1 between that classifier and one given only the timestamp.
  \item Trained for the CDD study's \matchedEpochs{} epochs, the study's fusion
  model raises macro-F1 relative to an image-only model on a random split of the
  AI Hub dataset, but at \fusionSeeds{} training runs per model, a test that
  holds out whole dates gives confidence intervals too wide to settle whether the
  gain exceeds zero on those dates.
\end{itemize}

\section{Materials and methods}
\label{sec:methods}

\subsection{Datasets}
\label{sec:corpora}

The Domestically Cultivated Subtropical and Tropical Pest and Disease Data
\citep{aihub2023subtropical}, the AI Hub dataset throughout this paper, is
distributed through AI Hub. Its metadata spreadsheet holds \numImages{} rows,
each naming one image file whose name encodes the image's crop, plant part, and
class. Each row also gives the farm, the date and time at which the image was
photographed, the time at which the row's panel was measured, and a soil and
ambient panel (air temperature, air humidity, soil
moisture, soil temperature, soil pH, nitrogen, phosphorus, potassium, and
electrical conductivity), although a row can lack some of the panel's values. The rows span \numClasses{} pest and disease classes
and a healthy class across \numCrops{} crops: apple mango, banana, dragon fruit,
fig, Hallabong mandarin, lemon, olive, and passion fruit. The dataset
also ships object-detection label files, in which each box marks one lesion or
pest with its class, but every analysis predicts one class per image and so takes each
image's class from its file name. The usage guideline that accompanies the dataset
\citep{aihub2023subtropical} records that the environmental sensors were installed per farm and that their readings were
matched to the images afterwards. The images were taken at \numFarms{} farms in
\photoWindow{}. Because a row's panel can be measured on another day than its
image was photographed, we take the day of a row's session to be the day on which
its panel was measured, so that all rows that carry one measurement of the panel
fall in one session.

The Crop Disease Diagnosis dataset \citep[CDD;][]{dacon2021cropdisease}, released
by LG AI Research through the Dacon challenge platform, pairs each of its
\cddImages{} labeled images of \cddCrops{} greenhouse crops with an annotation
file and an environment series. The annotation gives the date and time of the
photograph and the image's crop, disease, and disease severity, which together
form \cddClasses{} classes. The environment series is a table of greenhouse
sensor readings that the dataset presents as ending at the time the image was
taken.
The dataset stores each image under an identifier with its own environment CSV
file and annotation JSON file, so two images share an environment series only if
their CSV files hold the same readings.

Every analysis reads the files that the datasets ship, so every count and rate
that we report for the two datasets is measured on those files rather than taken
from their documentation. We use the
sensor values as the files ship them, including values implausible for their
variable, such as pH values outside the pH scale, because those are the values that a model trained on the dataset
receives. The diagnostic reads no
image: on the AI Hub dataset it reads the metadata spreadsheet, and on CDD the
annotation and environment files. Only the fusion experiment, which trains the
CDD study's fusion model on the AI Hub dataset, needs the imagery.

\subsection{The diagnostic}
\label{sec:diagnostic}

The diagnostic treats the session as the unit over which a sensor reading may be
shared. Our session corresponds to the sub-dataset of the Global Wheat Head Detection
benchmark, the images acquired over one experimental unit during one acquisition
session \citep{gwhd2021}: our farm or greenhouse corresponds to its experimental
unit, and our day to its acquisition session.

\subsubsection{Granularity probe}

The granularity probe establishes whether a dataset carries a reading per image or
shares each reading among the images of a farm or a session, whether a shared
reading identifies its session, and whether each sensor variable varies mostly
between farms. To test whether each AI Hub panel variable varies mostly between
farms, we computed the share of its variance that farm identity explains, which is $\eta^2$ of a one-way analysis of
variance with farm as the factor: the between-farm sum of squares divided by the
total sum of squares. Because sessions nest within farms, the share that
sessions explain is at least the share that farm explains, so a variable that
varies mostly between farms varies mostly between sessions as well. We call each
distinct set of a row's nine panel cells, an empty cell included, a panel tuple. We also counted the
panel tuples, the percentage of rows that share their panel tuple with at least
one other row, and the panel tuples that occur in only one session. To make the
corresponding counts on CDD, we had to decide when two images carry the same
environment series: CDD stamps each row of a series with its reading time, so the
files of two images can differ in their stamps alone. We therefore treated two images as carrying the same
environment series when their CSV files hold identical values once the column of
reading times is dropped. CDD names no greenhouse, so on CDD we approximate a session by the images that
share one series.
We counted the distinct series and the percentage of images that share their
series with another image. To test whether a series identifies one day and one
disease, we also computed two percentages over the series that more than one image
carries: the percentage whose images were all photographed on one date, and the
percentage whose images all carry one disease label, where an image's disease
label is the disease that its annotation records apart from the crop and the
severity. We read the counts by one rule: a dataset with one distinct reading per image
carries a per-image measurement, whereas a dataset with fewer distinct readings
than images shares its readings. For a dataset that shares its
readings, we take a variable to vary mostly between farms if farm identity
explains more than half of that variable's variance.

\subsubsection{Timestamp baseline}

The timestamp baseline uses random forests \citep{breiman2001random} from
scikit-learn \citep{pedregosa2011scikit}. One forest predicts each image's class
from the sensor readings, and another predicts it from the
timestamp alone (the timestamp classifier). On the AI Hub dataset, the timestamp is
the day and the time of day at which a row's panel was measured, so the timestamp
classifier reads the time of the same reading that the other forest reads.
The day enters as a count of calendar days, whose starting point does not change what a random forest
predicts, and the time of day enters in hours. On the AI Hub dataset, the forest that
reads the sensor readings, which we call the panel classifier, reads only the five
soil-chemistry variables of the panel: pH, N, P, K, and electrical conductivity.
It leaves out the other four panel variables because, within a farm, they change
with the time of day and the season, which the timestamp itself carries, so the comparison sets soil
chemistry against time. We trained and scored both classifiers within each crop,
because \numSingleCropClasses{} of the \numClasses{} pest and disease classes
occur on only one crop, so a classifier trained across crops could predict the
class by recognizing the crop. In a crop where one of the five takes a single value, the
classifier also leaves that variable out, because the variable carries no
information in that crop. A crop enters the comparison
of the two classifiers if at least \timestampMinRows{} of its rows carry all five
variables and those rows span at least two classes. Only the Hallabong mandarin fails this requirement,
because the dataset ships no potassium reading for it. The comparison therefore covers \timestampCrops{} of the dataset's
\numCrops{} crops. To bound the computation, we drew \timestampSampleRows{} of the rows that carry all
five variables at random within each crop. Every compared crop has more such rows
than that, so every crop supplies the full draw, and the threshold of
\timestampMinRows{} rows selects the same crops as any threshold up to the draw
size would. We scored every
classifier on the drawn rows by stratified \timestampFolds{}-fold cross-validation, which can
place rows of one session in both the training and the test folds, as the random
splits on which fusion gains are reported do. A random seed fixes the random draws of a run, which for the timestamp
baseline are the rows drawn, the folds, and the forest's random choices. We
repeated the draw and the cross-validation for each of \numSeeds{} random seeds,
so that every score in the comparison of the two classifiers carries a 95\%
confidence interval over seeds. Every classifier used the same
rows, folds, and random seed for its forest, so that the classifiers differ only
in their input columns. A majority-class
predictor, which always predicts the most frequent class of the training folds,
is scored on the same folds and gives the accuracy that a classifier reaches
without any input.

We scored accuracy and, as a check that no comparison depends on the sizes of
the classes, balanced accuracy. Balanced accuracy is the mean over classes of the
share of each class's rows that the classifier assigns to that class, so it gives
each class the same weight whatever its size. The timestamp classifier's margin in a
crop is its accuracy minus the panel classifier's. We call margins from
$-\comparableBand{}$ to $+\comparableBand{}$ the comparable band. The timestamp
classifier is comparable to or better than the panel classifier in a crop if its
margin, averaged over seeds, is at least $-\comparableBand{}$. Our timestamp criterion
is that the timestamp classifier is comparable to or better than the panel classifier
in a majority of crops. Meeting it would show that in most crops the soil-chemistry
readings predict the class no better than the timestamp, or better by at most the
half-width of the band, so that their accuracy gives no evidence that they do more
than identify the session. We also checked the criterion with no band, counting a
crop only if its mean margin is at least zero, so that the verdict of the criterion does not depend on the width of the band. The criterion uses seed means, so we also report the crops in which the
whole 95\% confidence interval of the margin lies outside the band: the timestamp
classifier is superior in such a crop if the interval lies above the band, and the
panel classifier is superior if the interval lies below the band.

A panel that identifies the session should predict little once the session is fixed,
so we also measured the panel classifier within sessions (the within-session
measurement). We scored the panel
classifier by stratified \timestampFolds{}-fold cross-validation inside each session
that holds at least \sessionMinRows{} rows and at least two classes of at least
\sessionMinClassRows{} rows each, after leaving out the rows of the classes below
that size, so that every fold holds each class
and enough rows to score an accuracy. In each of the \timestampCrops{} compared crops, all of
which hold such sessions, we averaged the classifier's accuracy over those
sessions, weighting each session by its number of scored rows. The panel classifier's advantage is its
accuracy minus that of the majority-class predictor. We computed this advantage
twice: within each session, and by one cross-validation over the scored rows of
all of the crop's scored sessions pooled together. The two values therefore differ only in whether the session is fixed. We report both values averaged over the crops and over the same \numSeeds{} random
seeds as the comparison of the two classifiers. We call the within-session advantage the
residual advantage. We fixed no threshold for calling the residual advantage
small, so we compare it with \comparableBand{}, the half-width of the comparable
band and the only scale that the timestamp criterion sets.

\subsection{Re-analysis of a published fusion study on CDD}
\label{sec:reanalysis}

The CDD study \citep{appliedsci2024dacon} scores its classifiers by macro-F1,
which is the unweighted mean of the F1 score over classes. The study reports that
adding the environment series to a ResNet-50 image classifier raises macro-F1 from \paperImageFone{} for the image model to \paperMultiFone{}
for the fusion model \citep[Table~2]{appliedsci2024dacon}, the gain of
\paperIncrement{} percentage points that its abstract quotes. The study obtained both macro-F1 values without its multimodal mixup
augmentation, which adds training examples that blend pairs of examples linearly
in both inputs. Our fusion experiment also leaves that augmentation out. The study's code is
public \citep{cropdiseaserepo}. Of the fusion studies we cite, this study is the
only one whose dataset and evaluation split were both released, so the experiment
behind its fusion claim can be re-run outside its author list.

We measured how much of the study's test set shares its environment series with
training. The study states a ratio of training, validation, and test images but
not how it drew its split, so we took the split from the study's repository,
which ships the split files themselves. The repository ships two sets of image
identifiers that each list the identifiers of a training, a validation, and a test
partition. The study's training script reads
one of the two sets, so the measurements below refer to that set, the study's
split, unless they name the other set, whose partitions hold the ratio that the
study states. We matched its identifiers
to the environment series and computed the leak rate, the share of test images
whose series also occurs in training. Dropping the column of reading times can only merge files into one series,
which can only raise the leak rate, so we also computed the leak rate on the raw
files with that column kept. We computed the leak rate
relative to the training partition and relative to the training images that
carry the test image's own label, because a series shared with an image of the
same class lets a classifier copy that class. We computed it relative to training
and validation together as well, because the study selects its model by the
validation score. We also counted the test identifiers that appear verbatim in
training. This count assumes nothing about the environment series and shows
whether the study tests its models on images that they were trained on.

To test whether a classifier predicts CDD's classes by recognizing environment
series that it saw in training, we drew two splits ourselves, each over \cddSeeds{} random seeds. Each seed divides the images into \cddFolds{} folds and holds out one fold for
testing. The folds are stratified by class alone (the random split) or also
grouped by environment series, so that each series, and so each CDD session, falls in exactly one fold (the
group-disjoint split). On the random and group-disjoint
splits, random forests with the settings of the timestamp baseline predict the \cddClasses{}-class label from
the environment series (the environment classifier), from the day and the time of
day of the photograph (the timestamp classifier), and from both. A majority-class
predictor is scored on the same test images. The environment features are the
mean, minimum, maximum, and first and last non-missing values of each value column of
the series. Missing values are imputed by the median of the training images. Our criterion for a classifier that recognizes series seen in training is that the
environment classifier loses more than \cddCollapseBand{} macro-F1 on the
group-disjoint split relative to the random split. Because the threshold is our
choice, we also report whether the environment classifier's interval on the
group-disjoint split lies entirely below its interval on the random split. A split by series can leave a
test image's photo date in training, so we also checked whether CDD admits a
split by date. We took each calendar month after the first as a test
block whose training data are the earlier months. In each crop whose images carry
at least two disease labels, counting healthy as one label, a test block kept the
test images whose disease label also occurs in the earlier months. We counted the
test blocks that keep images of at least two disease labels, because only such a
block can test whether a classifier separates diseases.

Finally, we ran the same random forests on the study's own split, with its
identifiers unchanged,
so their macro-F1 and the study's are measured over the same test images. We
ran them on the repository's other set of identifiers as well, to check whether a
result on the study's split holds on the other set that the repository ships. The
forests tune nothing, so they train on the study's training and validation
partitions together. Our criterion on the study's split is that the mean macro-F1
of the timestamp and the environment features together is at least the macro-F1
that the study reports for its fusion model, which would show
that inputs that identify the session can account for the published score.
Beside the criterion, we report whether the macro-F1 that the study reports for
its fusion model lies inside the interval of the timestamp classifier alone, in
which case a timestamp alone matches the published score. The
study's split is fixed, so our intervals on it carry only the
variance of the forest over \cddTheirSeeds{} random seeds, just as the intervals
that the study reports carry only the variance that the random seed of its
training introduces.

\subsection{Image--sensor fusion experiment on the AI Hub dataset}
\label{sec:fusion-design}

The fusion experiment asks whether the fusion model of the CDD study produces a
gain on the AI Hub dataset, and whether the gain still exceeds zero on a split that holds out
time. Only a dataset in which a date-disjoint split leaves every class on both
sides can test whether the gain exceeds zero on such a split. A date-disjoint split of the AI Hub
dataset keeps every class on both sides, whereas no date-disjoint split of CDD can,
because \cddSingleDateClasses{} of its \cddClasses{} classes were photographed on a single
date (Section~\ref{sec:res-cdd-group}).

We compared \gateAxes{} candidate splits to choose the unit
that the experiment holds out. The candidates were a random split and splits
disjoint by farm, by photo date, and by the pair of farm and photo date. For
\pctPhotoDayDiffers{}\% of the rows, the image was photographed on another day than
the one on which the row's panel was measured, so a split by photo date can place
rows that share one session's panel in both the training and the test data. For each split, we divided the whole metadata spreadsheet into \gateFolds{}
folds, at random or by the split's unit, and measured two quantities averaged over
the folds. The leak rate, defined as on CDD, is the share of test images whose panel tuple
recurs in training, so a high leak rate lets a model recognize a test image's
session from its panel. Coverage is the share of test images whose combination of crop, plant
part (the organ photographed), and class also occurs in training. A split low in
coverage measures how the models handle unseen combinations rather than what the
panel contributes, so we set a floor of \positivityFloor{}\% on
coverage. A
random split has a leak rate of \gateRandomLeak{}\%, so it leaves the panel tuple of
nearly every test image in training. A farm-disjoint split has a
leak rate of zero but covers only \gateFarmCellPos{}\% of its test images, below that floor.
The splits disjoint by photo date and by the pair of farm and photo date have
leak rates of \gateDateLeak{}\% and \gateSessionLeak{}\% and cover every test image, so any floor
above the farm-disjoint split's coverage keeps these two splits and excludes the
farm-disjoint one. The experiment
holds out photo dates. The splits by photo date and by the pair of farm and photo
date both keep together the images of each farm and photo date, but only the split
by photo date also keeps each test image's photo date out of training.

The fusion experiment trains an image-only model and a fusion model on a
subsample of the AI Hub dataset, which we drew to bound the training cost:
\subsampleSize{} images stratified by crop, plant part, and class. Besides the
date-disjoint split, we kept a farm-disjoint split although its coverage falls
below the floor, because the date-disjoint split still shares farms between its
training and test images. We split the subsample three ways, each holding out one
of \fusionFolds{} folds for testing: at random (the in-distribution split), by
photo date (date-disjoint), and by farm (farm-disjoint). The test images of the
in-distribution split share farms and dates with its training images. We report
the farm-disjoint gain beside its coverage. A
validation set was drawn from the training images by the same grouping as the
test images, to choose the epoch whose weights were kept. An epoch is one pass
over the training images.  The image-only model is a
ResNet-50 \citep{he2016deep} pretrained on ImageNet \citep{deng2009imagenet},
taken from the timm library \citep{wightman2019timm}. The fusion model is a late
fusion, which combines the class scores (logits) of the ResNet-50 with those of
an environmental branch in a fusion layer. The branch and the fusion layer are
the CDD study's own, taken from the model file that its repository releases
\citep{cropdiseaserepo}. We resized the output layers of the branch and of the
fusion layer to this dataset's classes. The branch is a long short-term memory (LSTM) network
\citep{hochreiter1997lstm}, a recurrent network for sequences, followed by two
fully connected layers. The fusion layer randomly zeroes a share of the two sets
of class scores during training (dropout), joins them into one vector, and maps
that vector to the classes with a fully connected layer. The branch was written
for a time series, whereas each AI Hub image carries a single set of panel values, so the branch
reads a sequence of length one. Unlike the panel classifier, the branch reads all nine panel variables, because
the experiment tests the panel as a fusion model receives it.
A missing value is set to zero, as in the CDD study's code. Each variable is
then scaled to the unit interval by its minimum and maximum over the training
images.

Both models follow the CDD study's training procedure
in its augmentation of the training images by random transformations, its
sampling of training images in inverse proportion to their class frequency, its
Adam optimizer \citep{kingma2015adam}, its schedule for the optimizer's learning
rate, and its selection of the weights from the epoch with the best validation
macro-F1. The learning rate is the step size of the optimizer's weight updates.
The schedule, cosine annealing with warm restarts, lowers the rate along a
cosine curve and returns it to its initial value at fixed intervals
\citep{loshchilov2017sgdr}. To bound the computation, we first trained both models
on all three splits at a reduced budget that departs from the study's procedure in
three settings: \lowBudgetEpochs{} epochs, batches of \lowBudgetBatch{} images,
and early stopping, which ends a run once the validation score stops improving.
We then trained both models on
the in-distribution and date-disjoint splits at the matched budget, the study's
own \matchedEpochs{} epochs with batches of \matchedBatch{} images and no early
stopping, because the interval of the in-distribution gain included zero at the
reduced budget and a reduced budget can undertrain the environmental branch. We
did not retrain on the farm-disjoint split at the matched budget, because its
coverage falls below the floor and its gain therefore measures in part how the models handle unseen
combinations of crop, plant part, and class at any budget.

The image-only and fusion models share the ResNet-50, the training procedure, and
the random seeds, so that they differ only in the panel and in the layers that
read it. The fusion gain
\begin{equation}
\delta = \mathrm{macroF1}(\text{image}+\text{panel}) - \mathrm{macroF1}(\text{image})
\label{eq:gain}
\end{equation}
is therefore what the panel contributes to macro-F1 through those layers, taken
within a seed. Every seed trains on the same split, so the 95\% confidence intervals of the
gains carry only the variance of training,
such as that of the initial weights of the layers that are not pretrained and of
the order of the training images. The study's finding, that fusion raises macro-F1, replicates on
this dataset if the gain's 95\% confidence interval excludes zero, so replication in this experiment concerns whether a gain exists and not its size. The paired reduction is the
in-distribution gain minus the date-disjoint gain, taken within each seed. A
replicated gain collapses under the date-disjoint split if the date-disjoint
gain's interval includes zero and the paired reduction's interval excludes zero.
A replicated gain remains above zero on that split if the date-disjoint gain's
interval excludes zero.

\subsection{Statistical analysis and software}
\label{sec:stats}

We ran each comparison on a fixed set of random seeds
(Table~\ref{tab:settings}). We report
each score from a set of seeds as the mean over those seeds with a 95\% confidence interval from the $t$
distribution. Where two models are scored on the same test images, we take their
difference within seed and conclude that one score exceeds the other only where the
interval of the difference excludes zero. A score on one split, whose test images
differ from those of another split, is lower than a score on the other split only
where its interval lies entirely below the interval of that score. For the CDD
classifiers and the fusion models, we report balanced accuracy beside macro-F1,
whereas the criteria read macro-F1, the CDD study's metric.

\begin{table}[tbp]
\centering
\caption{Settings of the classifiers and of the fusion training. A random-forest setting that neither this table nor the text
gives is the scikit-learn default, and a fusion-training setting that neither
gives is the CDD study's.}
\label{tab:settings}
\begin{tabular}{>{\raggedright\arraybackslash}p{0.42\textwidth}>{\raggedright\arraybackslash}p{0.48\textwidth}}
\toprule
Setting & Value \\
\midrule
\multicolumn{2}{l}{\emph{Timestamp baseline and CDD re-analysis}} \\
Random forest & \forestTrees{} trees \\
AI Hub rows drawn per crop and seed & \timestampSampleRows{} \\
AI Hub cross-validation & stratified, \timestampFolds{} folds \\
AI Hub seeds & \numSeeds{} \\
CDD test partition (drawn splits) & one of \cddFolds{} folds \\
CDD seeds, drawn splits & \cddSeeds{} \\
CDD seeds, the study's split & \cddTheirSeeds{} \\
\midrule
\multicolumn{2}{l}{\emph{Fusion experiment on the AI Hub dataset}} \\
Subsample & \subsampleSize{} images \\
Test partition & one of \fusionFolds{} folds \\
Image size & \imagePixels{} $\times$ \imagePixels{} pixels \\
Environmental branch & LSTM, \lstmHidden{} hidden units \\
Optimizer & Adam, learning rate \learningRate{} \\
Learning-rate schedule & cosine annealing with warm restarts every \restartEpochs{} epochs \\
Reduced budget & \lowBudgetEpochs{} epochs, batches of \lowBudgetBatch{} images,
early stopping after \earlyStopPatience{} epochs without improvement \\
Matched budget & \matchedEpochs{} epochs, batches of \matchedBatch{} images, no
early stopping \\
Seeds & \fusionSeeds{} \\
\bottomrule
\end{tabular}
\end{table}

Five parts of this study were decided after we had seen the data, so what
rests on them is exploratory rather than confirmatory: our assessment of whether
the fusion studies that the Introduction cites compare their sensor readings with
a timestamp and report a farm- or session-disjoint split, the decision to count distinct sensor readings, the
timestamp criterion with its comparable band, the matched-budget training, and the
recommendations of the Discussion. The results that rest on these parts are the
counts of distinct readings, the fusion results at the matched budget, and every
comparison that uses those counts or the comparable band. These comparisons are
the timestamp comparison on the AI Hub dataset, the within-session measurement, and
the comparison of the random and group-disjoint splits on CDD, whose
group-disjoint split groups the images by distinct environment series. By
contrast, the criterion on the CDD study's split and the criteria of the fusion
experiment do not depend on the five parts, although the fusion experiment's
results at the matched budget do. The leak rates of the two sets of identifiers
that the CDD study's repository ships and of the AI Hub splits do not depend on
the five parts either, because a leak rate is a share of test images, whereas the
counts of distinct readings and the group-disjoint split take each distinct
reading as a unit. Nor does any rule of ours for matching sensor values change
these leak rates: on CDD, matching the raw files gives the same rates on both sets
of identifiers, because nearly
every series that images share is stored in identical files, stamps included;
on the AI Hub dataset, a leak rate needs no matching rule, because a panel tuple
holds a row's nine cells as the spreadsheet records them, an empty cell included.

The analyses ran in Python \pythonVersion{} with pandas \pandasVersion{},
scikit-learn \sklearnVersion{}, PyTorch \torchVersion{}
\citep{paszke2019pytorch}, timm \timmVersion{}, and Albumentations
\albumentationsVersion{} \citep{buslaev2020albumentations}. The analysis scripts
were written with the assistance of large language models (Claude Opus 4.8 and
Claude Opus 5.5, Anthropic) from requirements and pseudocode that the author
specified. The author reviewed and verified the scripts.

\section{Results}
\label{sec:results}

\subsection{One sensor reading is shared by many images in both datasets}
\label{sec:res-diagnostic}

\begin{figure}[tbp]
\centering
\includegraphics[width=\textwidth]{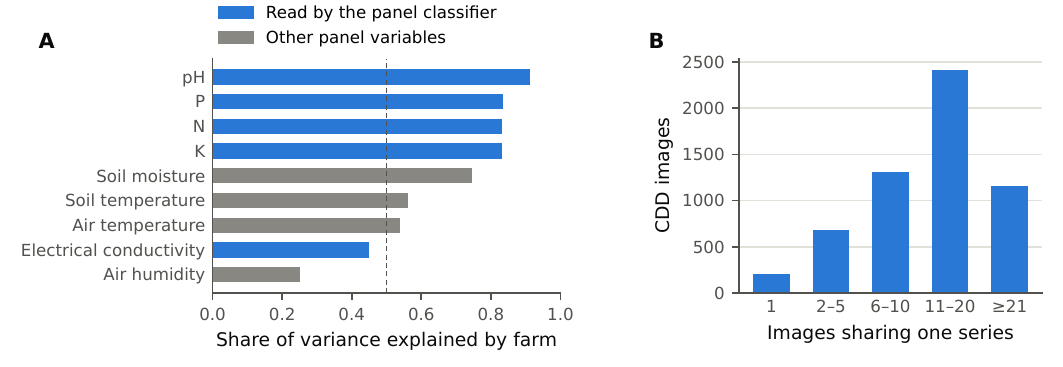}
\caption{Farm identity explains more than half the variance of every panel
variable of the AI Hub dataset except air humidity and electrical conductivity,
and many CDD images share each environment series.
\textbf{(A)} The share of each panel variable's variance that farm identity
explains on the AI Hub dataset. The dashed line marks one half. The five soil-chemistry
variables that the panel classifier reads are shown in blue.
\textbf{(B)} The CDD images grouped by how many images share their environment
series. A series shared by one image is a per-image reading.}
\label{fig:granularity}
\end{figure}

On both datasets, one sensor reading is shared across many images, so the
readings describe groups of images rather than measuring each one. On the AI Hub dataset, farm identity explains
more than half the variance of every panel variable except air humidity and
electrical conductivity (Figure~\ref{fig:granularity}A). Farm identity explains
between \rsqFarmLo{} and \rsqFarmHi{} of the variance of pH, N, P, and K. The dataset carries only
\numPanelTuples{} distinct panel tuples across its \numImages{} metadata rows. Of
the rows, \pctPanelTupleShared{}\% share their panel tuple with at least one other
row. Of the panel tuples, \numTuplesOneSession{} occur in only one session, so
nearly every panel tuple identifies its session.

The classifiers and the fusion model receive the AI Hub panel values uncorrected,
although many are implausible as soil readings: \pctPhImpossible{}\% of the pH
values fall outside $[0,14]$, the range of the pH scale, and \pctNpkZero{}\% of the
recorded N, P, and K values are exactly zero.

The granularity probe finds the same sharing on CDD. Only \cddSeries{} distinct
environment series exist for its \cddImages{} images. Of the
images, \cddSharedPct{}\% share their series with at least one other image
(Figure~\ref{fig:granularity}B). Of the shared series, \cddSingleDatePct{}\% span a
single photo date, so nearly every shared series identifies one day. Of the shared series, \cddSingleCodePct{}\% carry a single disease label, so
nearly every shared series identifies one disease.

\subsection{A timestamp is comparable to or better than the soil-chemistry panel in every crop compared}
\label{sec:res-timestamp}

\begin{figure}[tbp]
\centering
\includegraphics[width=0.5\textwidth]{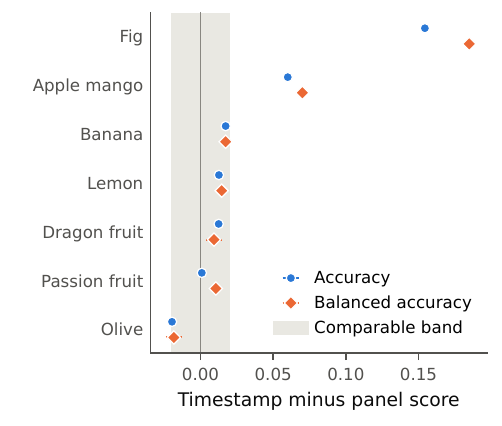}
\caption{Timestamp classifier minus panel classifier on the AI Hub dataset, per crop,
in accuracy and in balanced accuracy. Markers are means over \numSeeds{} random
seeds, and bars are 95\% confidence intervals; a marker hides an interval narrower
than itself.
The shaded region is the comparable band, within which the timestamp classifier is
comparable to the panel classifier. Values are positive where the timestamp classifier is the more accurate.}
\label{fig:timestamp}
\end{figure}

On the AI Hub dataset, the timestamp classifier is comparable to or better than the
panel classifier in \timestampGeqCrops{} of \timestampCrops{} crops over \numSeeds{}
random seeds, and the panel classifier is superior in \panelDecisiveCrops{} of them
(Figure~\ref{fig:timestamp}; Table~\ref{tab:timestamp}). The timestamp classifier is
superior in fig and apple mango, whose intervals lie above the comparable band.
Olive, the only crop with a negative margin, still counts as comparable, because
its mean margin, $\oliveLeanPlainFine{}$, lies just inside the band. The interval
of olive's margin, $\oliveLeanCI{}$, excludes zero, so the panel classifier scores
higher in olive, but the interval does not lie below the band, so the panel
classifier is not superior there. Because no
other crop has a negative margin, the check with no band counts every crop but
olive, and those crops are still a majority, so the timestamp criterion also holds with no band. The
per-crop comparisons are the same under balanced accuracy: the margins of fig and
apple mango grow, and olive stays within the band. Both classifiers exceed the majority-class predictor scored on the same
folds (Table~\ref{tab:timestamp}).

\begin{table}[tbp]
\centering
\caption{Timestamp classifier versus panel classifier at disease-class prediction on
the AI Hub dataset. Values are means over \numSeeds{} random seeds with 95\%
confidence intervals, and the first three rows are also means over crops. The
per-crop rows give the timestamp classifier's margin, its score minus the
panel classifier's in the column's metric, for the two crops in which the timestamp classifier is superior and for
the one crop with a negative margin. Figure~\ref{fig:timestamp} shows the margins of all \timestampCrops{} crops.}
\label{tab:timestamp}
\begin{tabular}{lcc}
\toprule
 & Accuracy & Balanced accuracy \\
\midrule
Majority class, crop mean & $\basePooledPlain{}$~~$\basePooledPlainCI{}$ & $\basePooledBA{}$~~$\basePooledBACI{}$ \\
Panel, crop mean          & $\panelPooledPlain{}$~~$\panelPooledPlainCI{}$ & $\panelPooledBA{}$~~$\panelPooledBACI{}$ \\
Timestamp, crop mean          & $\timestampPooledPlain{}$~~$\timestampPooledPlainCI{}$ & $\timestampPooledBA{}$~~$\timestampPooledBACI{}$ \\
\midrule
\multicolumn{3}{l}{Timestamp margin per crop} \\
\quad Fig         & $\figWinPlain{}$~~$\figWinCI{}$       & $\figWinBA{}$~~$\figWinBACI{}$ \\
\quad Apple mango & $\mangoWinPlain{}$~~$\mangoWinCI{}$   & $\mangoWinBA{}$~~$\mangoWinBACI{}$ \\
\quad Olive       & $\oliveLeanPlain{}$~~$\oliveLeanCI{}$ & $\oliveLeanBA{}$~~$\oliveLeanBACI{}$ \\
\bottomrule
\end{tabular}
\end{table}

Averaged over the crops, the panel classifier's advantage falls most of the way to
zero once the session is fixed, as it should if the panel identifies the session: it is
\panelAdvantagePooled{} over the scored rows of the sessions pooled and
\panelAdvantageWithinSession{} within single sessions, where the date is fixed, and
the fall, paired within seed, has an interval that excludes zero. Both values
come from the rows of the sessions that the within-session measurement scores,
whereas Table~\ref{tab:timestamp} scores all drawn rows of each crop, so the
pooled advantage differs from the panel classifier's crop mean minus the
majority-class predictor's in Table~\ref{tab:timestamp}. The residual advantage stays above zero, however: it ranges from \residualLo{} to
\residualHi{} over the crops, and in every crop its interval lies above
\comparableBand{}, the half-width of the comparable band.

\subsection{The published CDD split shares environment series between training and test data, and a timestamp matches the reported fusion score on it}
\label{sec:res-cdd}

The study's split leaks. For \cddLeak{}\% of its test images, the identical
environment series also occurs in training (\cddLeakMapped{}\% with the test
image's own label, \cddLeakTrainval{}\% in training and validation together).
Of the \cddGroupsTotal{} distinct test series, \cddGroupsShared{} recur in
training. Matching the raw files gives the same leak rate on the study's split. The repository's other set of identifiers leaks \cddSecondSplitLeak{}\% into its
training partition, so neither shipped split is group-disjoint on the environment series. A random
split that we draw leaks \cddLeakRandom{}\%, and a group-disjoint one none
(Table~\ref{tab:cdd}). The study's identifiers also leak at the level of images.
Of its test
identifiers, \cddDirectTest{}\% appear verbatim in its training partition and
\cddDirectTrainval{}\% in its training and validation partitions together, so the
study scores its models partly on images that they were trained on.

\begin{table}[tbp]
\centering
\caption{Macro-F1 and balanced accuracy of the CDD predictors on the two splits
that we draw over the whole dataset, one random and one disjoint on the environment series, each over
\cddSeeds{} random seeds with 95\% confidence intervals. Each row is a random
forest on the features that the row names, or the constant majority-class
predictor. No row uses the images or the study's split. Paired differences
are taken within seed. The leak rate is the share of test images whose
environment series recurs in training.}
\label{tab:cdd}
\begin{tabular}{>{\raggedright\arraybackslash}p{0.30\textwidth}cc}
\toprule
 & Random split & Group-disjoint split \\
\midrule
\multicolumn{3}{l}{\emph{Macro-F1}} \\
Majority class & $\cddBaseRandom{}$~~$\cddBaseRandomCI{}$ & $\cddBaseGroup{}$~~$\cddBaseGroupCI{}$ \\
Environment & $\cddEnvRandom{}$~~$\cddEnvRandomCI{}$ & $\cddEnvGroup{}$~~$\cddEnvGroupCI{}$ \\
Timestamp & $\cddTimestampRandom{}$~~$\cddTimestampRandomCI{}$ & $\cddTimestampGroup{}$~~$\cddTimestampGroupCI{}$ \\
Timestamp and environment & $\cddTimestampEnvRandom{}$~~$\cddTimestampEnvRandomCI{}$ & $\cddTimestampEnvGroup{}$~~$\cddTimestampEnvGroupCI{}$ \\
\midrule
Timestamp and environment minus timestamp & $\cddTimestampEnvOverTimestamp{}$~~$\cddTimestampEnvOverTimestampCI{}$ & \\
Timestamp and environment minus environment & $\cddTimestampEnvOverEnv{}$~~$\cddTimestampEnvOverEnvCI{}$ & \\
Timestamp minus environment & $\cddTimestampOverEnvRandom{}$~~$\cddTimestampOverEnvRandomCI{}$ & $\cddTimestampOverEnvGroup{}$~~$\cddTimestampOverEnvGroupCI{}$ \\
\midrule
\multicolumn{3}{l}{\emph{Balanced accuracy}} \\
Majority class & $\cddBaseRandomBA{}$~~$\cddBaseRandomBACI{}$ & $\cddBaseGroupBA{}$~~$\cddBaseGroupBACI{}$ \\
Environment & $\cddEnvRandomBA{}$~~$\cddEnvRandomBACI{}$ & $\cddEnvGroupBA{}$~~$\cddEnvGroupBACI{}$ \\
Timestamp & $\cddTimestampRandomBA{}$~~$\cddTimestampRandomBACI{}$ & $\cddTimestampGroupBA{}$~~$\cddTimestampGroupBACI{}$ \\
Timestamp and environment & $\cddTimestampEnvRandomBA{}$~~$\cddTimestampEnvRandomBACI{}$ & $\cddTimestampEnvGroupBA{}$~~$\cddTimestampEnvGroupBACI{}$ \\
Timestamp minus environment & $\cddTimestampOverEnvRandomBA{}$~~$\cddTimestampOverEnvRandomBACI{}$ & $\cddTimestampOverEnvGroupBA{}$~~$\cddTimestampOverEnvGroupBACI{}$ \\
\midrule
Leak rate (\%) & $\cddLeakRandom{}$ & $\cddLeakGroup{}$ \\
\bottomrule
\end{tabular}
\end{table}

\begin{figure}[tbp]
\centering
\includegraphics[width=\textwidth]{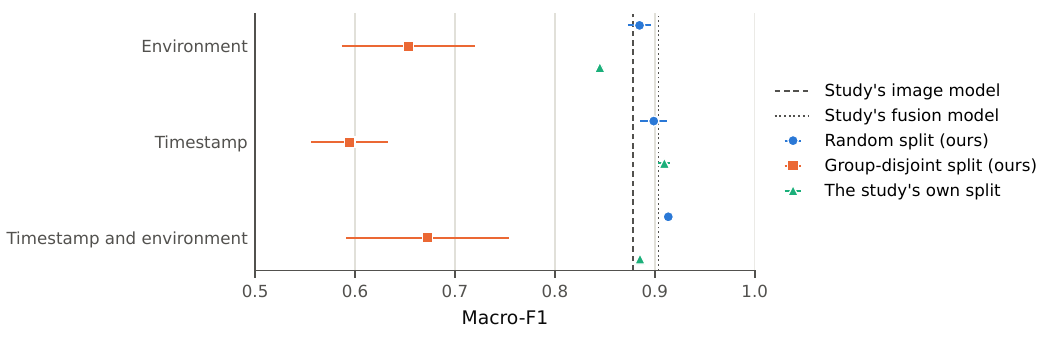}
\caption{Macro-F1 of three CDD predictors on the random and group-disjoint splits
that we draw and on the study's own split. Markers are means and bars are 95\%
confidence intervals. The vertical lines mark the macro-F1 that the study reports
for its image model and for its fusion model on its own split.
The majority-class predictor is omitted; Tables~\ref{tab:cdd}
and~\ref{tab:cdd-theirs} give its scores.}
\label{fig:cdd}
\end{figure}

On the study's own split, the timestamp and the environment features together
reach \cddTheirTimestampEnv{}, above the \paperImageFone{} that the study reports for its
image model but short of the \paperMultiFone{} that it reports for its fusion model
(Table~\ref{tab:cdd-theirs}). The
timestamp classifier, however, reaches \cddTheirTimestamp{} although it reads
nothing but the day and the time of day of the photograph, so a timestamp alone matches the macro-F1
that the study reports for its fusion model (Figure~\ref{fig:cdd}). The timestamp alone exceeds the timestamp and environment features together in a
paired difference whose interval excludes zero, so adding the environment features
to the timestamp lowers the forest's macro-F1 on this split. The timestamp alone also exceeds the
timestamp and environment features together on the repository's other set of
identifiers, which the study's training script does not read. On the study's split, the environment features alone nonetheless reach \cddTheirEnv{}, far above the
\cddTheirBase{} of the majority-class predictor.

\begin{table}[tbp]
\centering
\caption{The predictors of Table~\ref{tab:cdd} scored on the two sets of
identifiers in the CDD study's repository that hold a test partition: the study's split, from which its
training script reads the training, validation, and test partitions, and the other
set. Each split is fixed, so the
\cddTheirSeeds{} random seeds behind each interval vary the forest alone, just as the runs behind the intervals that the study reports vary only the
random seed of its training. Values
are given to four decimals, the precision of the study's reported figures. The
paired difference is taken within seed. No row uses the images.}
\label{tab:cdd-theirs}
\begin{tabular}{>{\raggedright\arraybackslash}p{0.30\textwidth}cc}
\toprule
 & The study's split & The other set \\
\midrule
\multicolumn{3}{l}{\emph{Macro-F1}} \\
Majority class & $\cddTheirBase{}$~~$\cddTheirBaseCI{}$ & $\cddOtherBase{}$~~$\cddOtherBaseCI{}$ \\
Environment & $\cddTheirEnv{}$~~$\cddTheirEnvCI{}$ & $\cddOtherEnv{}$~~$\cddOtherEnvCI{}$ \\
Timestamp & $\cddTheirTimestamp{}$~~$\cddTheirTimestampCI{}$ & $\cddOtherTimestamp{}$~~$\cddOtherTimestampCI{}$ \\
Timestamp and environment & $\cddTheirTimestampEnv{}$~~$\cddTheirTimestampEnvCI{}$ & $\cddOtherTimestampEnv{}$~~$\cddOtherTimestampEnvCI{}$ \\
\midrule
Timestamp minus timestamp and environment & $\cddTheirTimestampOverTimestampEnv{}$~~$\cddTheirTimestampOverTimestampEnvCI{}$ & $\cddOtherTimestampOverTimestampEnv{}$~~$\cddOtherTimestampOverTimestampEnvCI{}$ \\
\midrule
\multicolumn{3}{l}{\emph{Balanced accuracy}} \\
Majority class & $\cddTheirBaseBA{}$~~$\cddTheirBaseBACI{}$ & $\cddOtherBaseBA{}$~~$\cddOtherBaseBACI{}$ \\
Environment & $\cddTheirEnvBA{}$~~$\cddTheirEnvBACI{}$ & $\cddOtherEnvBA{}$~~$\cddOtherEnvBACI{}$ \\
Timestamp & $\cddTheirTimestampBA{}$~~$\cddTheirTimestampBACI{}$ & $\cddOtherTimestampBA{}$~~$\cddOtherTimestampBACI{}$ \\
Timestamp and environment & $\cddTheirTimestampEnvBA{}$~~$\cddTheirTimestampEnvBACI{}$ & $\cddOtherTimestampEnvBA{}$~~$\cddOtherTimestampEnvBACI{}$ \\
Timestamp minus timestamp and environment & $\cddTheirTimestampOverTimestampEnvBA{}$~~$\cddTheirTimestampOverTimestampEnvBACI{}$ & $\cddOtherTimestampOverTimestampEnvBA{}$~~$\cddOtherTimestampOverTimestampEnvBACI{}$ \\
\midrule
Leak rate into training and validation (\%) & $\cddLeakTrainval{}$ & $\cddOtherLeakTrainval{}$ \\
\bottomrule
\end{tabular}
\end{table}

\subsection{Much of the CDD environment classifier's random-split score comes from recognizing series seen in training}
\label{sec:res-cdd-group}

On the group-disjoint
split, both the environment classifier and
the timestamp classifier score a lower macro-F1 than on the random split, each with an interval
that lies entirely below its random-split interval (Table~\ref{tab:cdd}), so part of
each classifier's random-split score depends on series shared between training
and test images. The environment classifier falls from \cddEnvRandom{} to
\cddEnvGroup{}, a drop larger than the \cddCollapseBand{} that our criterion sets, so much of
the environment classifier's random-split score comes from recognizing series
that it saw in training.

Neither classifier falls to the \cddBaseGroup{} that the majority-class predictor
scores on the group-disjoint split. Photo date, which spans several series, is a confound that the group-disjoint
split leaves in place: it can explain the scores that both classifiers keep on
that split if series recorded on one date resemble each other. The \cddSeries{} series spread over \cddDates{} photo dates,
\cddSeriesPerDate{} to a date, so a held-out series leaves its own date in
training. Holding out whole dates would remove this confound, but no split by photo
date keeps every class of CDD on both sides, because \cddSingleDateClasses{} of its
\cddClasses{} classes were photographed on a single date. A split forward in time
cannot test whether a classifier separates diseases either: when each calendar month
serves as a test block for the months before it, only one test block keeps images of
at least two disease labels, and that block keeps only \cddTransferableImages{}
images.

In macro-F1, the timestamp baseline detects no difference between the timestamp and the
environment series on the two splits that we draw over the whole of CDD, one that
shares series between training and test images and one that does not: the timestamp
classifier's macro-F1 minus the environment classifier's is
$\cddTimestampOverEnvRandom{}$ $\cddTimestampOverEnvRandomCI{}$ on the random split and
$\cddTimestampOverEnvGroup{}$ $\cddTimestampOverEnvGroupCI{}$ on the group-disjoint
split, both with intervals that include zero (Table~\ref{tab:cdd}). In balanced
accuracy, the same difference on the group-disjoint split is
$\cddTimestampOverEnvGroupBA{}$ $\cddTimestampOverEnvGroupBACI{}$, so there the
environment classifier scores higher.
Under the random split, however, the timestamp and the environment features together
exceed either alone, in paired differences whose intervals exclude zero, although
on the study's split the combination scored below the timestamp alone
(Table~\ref{tab:cdd-theirs}). Combining two inputs that both identify the session can
therefore raise macro-F1 above either input alone on a split that shares series,
so a fusion gain on such a split does not by itself show that the sensor readings
add agronomic information to the image.

\subsection{Late fusion raises macro-F1 in distribution on the AI Hub dataset, and the date-disjoint test is inconclusive in macro-F1}
\label{sec:fusion}

In macro-F1, the in-distribution fusion gain replicates at the matched budget but
not at the reduced budget (Table~\ref{tab:fusion}). Relative to an image-only macro-F1 of
\imageOnlyFoneMatched{}, the matched-budget gain is $\delta = \iidDelta{}$, with an interval that
excludes zero. At the reduced budget, the in-distribution gain is
$\delta = \iidDeltaLowBudget{}$ relative to an image-only macro-F1 of
\imageOnlyFone{}, with an interval that includes zero. Both models
score lower at the matched budget than at the reduced budget in every seed: the
image-only model's macro-F1 changes by $\imageOnlyBudgetChange{}$ and the fusion
model's by $\fusionBudgetChange{}$, both with intervals that exclude zero, so the gain widens
because the image-only model loses twice as much as the fusion model, not because the fusion model improves.

\begin{table}[tbp]
\centering
\caption{Late-fusion gain $\delta$ on the AI Hub dataset in macro-F1 and in balanced
accuracy, the fusion model's score minus the image-only model's, with the
image-only model's macro-F1 beside it. Values are means over \fusionSeeds{} random seeds, with 95\%
confidence intervals for $\delta$. The reduction is the in-distribution gain
minus the date-disjoint gain, paired within seed. The farm-disjoint test fold
falls below the coverage floor of Section~\ref{sec:fusion-design}.}
\label{tab:fusion}
\begin{tabular}{lccc}
\toprule
Split and budget & Image-only macro-F1 & $\delta$ & 95\% CI \\
\midrule
\multicolumn{4}{l}{\emph{Macro-F1}} \\
In distribution, reduced budget & $\imageOnlyFone{}$ & $\iidDeltaLowBudget{}$ & $\iidDeltaLowBudgetCI{}$ \\
Date-disjoint, reduced budget & $\imageOnlyFoneDate{}$ & $\dateDeltaLowBudget{}$ & $\dateDeltaLowBudgetCI{}$ \\
Farm-disjoint, reduced budget & $\imageOnlyFoneFarm{}$ & $\farmDeltaLowBudget{}$ & $\farmDeltaLowBudgetCI{}$ \\
In distribution, matched budget & $\imageOnlyFoneMatched{}$ & $\iidDelta{}$ & $\iidDeltaCI{}$ \\
Date-disjoint, matched budget & $\imageOnlyFoneDateMatched{}$ & $\dateDelta{}$ & $\dateDeltaCI{}$ \\
Reduction, matched budget &  & $\dateReduction{}$ & $\dateReductionCI{}$ \\
\midrule
\multicolumn{4}{l}{\emph{Balanced accuracy}} \\
In distribution, reduced budget & & $\iidDeltaLowBudgetBA{}$ & $\iidDeltaLowBudgetBACI{}$ \\
Date-disjoint, reduced budget & & $\dateDeltaLowBudgetBA{}$ & $\dateDeltaLowBudgetBACI{}$ \\
Farm-disjoint, reduced budget & & $\farmDeltaLowBudgetBA{}$ & $\farmDeltaLowBudgetBACI{}$ \\
In distribution, matched budget & & $\iidDeltaBA{}$ & $\iidDeltaBACI{}$ \\
Date-disjoint, matched budget & & $\dateDeltaBA{}$ & $\dateDeltaBACI{}$ \\
Reduction, matched budget & & $\dateReductionBA{}$ & $\dateReductionBACI{}$ \\
\bottomrule
\end{tabular}
\end{table}

In macro-F1, every gain on the splits that hold out dates or farms has an interval
that includes zero (Table~\ref{tab:fusion}). In balanced accuracy, by contrast, every
gain except the farm-disjoint one has an interval that excludes zero, so in that
metric the gain remains above zero on held-out dates at both budgets. On the subsample, the in-distribution split has a leak rate of \iidPanelLeak{}\% and
the date-disjoint split a leak rate of \datePanelLeak{}\%, so the
date-disjoint split removes most of what lets a model recognize a test image's
session from its panel. At the matched budget,
the paired reduction also has an interval that includes zero, so the
matched-budget training on the date-disjoint split establishes in macro-F1 neither
that the gain collapses nor that it remains above zero. Coverage differs between the
folds of the farm-disjoint split, which cover \farmCellPosMean{}\% of their test
images on average: the test fold on which we scored the models covers only
\farmCellPos{}\%, below the \positivityFloor{}\% floor, so
the farm-disjoint gain is not an estimate of what the panel contributes.

\section{Discussion}
\label{sec:discussion}

\subsection{The sensor readings of both datasets identify the session}

The readings that the two datasets attach to their images mainly identify the session
rather than measure each plant's environment. Most of the predictive
value of the AI Hub dataset's soil-chemistry readings is farm and session
identity, because restricting prediction to one session removes most of the
panel classifier's advantage on average across crops. A timestamp is also
comparable to or better than those readings at predicting the class, so their
accuracy over a whole crop gives no evidence that they do more than identify the
session. Within a session, by contrast, the residual advantage exceeds the
half-width of the comparable band in every crop, so the soil-chemistry readings still
separate classes within a session. The readings can separate classes within a
session without measuring the plant's environment, however, because a reading
within a session can be attached to the images taken until the next reading and
so identify the part of the day in which they were taken.

Because the readings mainly identify the session, we expect a fusion model that
relies on them to lose the part of its gain that comes from session identity when
it is tested on farms or dates outside its training data, provided that the classes
differ between sessions. Image classifiers for plant disease likewise lose accuracy
on images taken under other conditions \citep{mohanty2016deep,wu2023lab2field}.

\subsection{What the re-analysis of a published fusion gain can show}

On the published CDD split, a timestamp alone reaches the macro-F1 that the CDD
study \citep{appliedsci2024dacon} reports for its fusion model, so nothing in the numbers that the study
publishes separates session
identity from an agronomic contribution of the environment series. A reader
cannot detect the leak from the study's own confidence intervals, because the
runs behind them
vary only the random seed of training over fixed split files \citep{appliedsci2024dacon}
and a leak raises the score rather than its spread. A re-analysis can expose such
a leak only where a study releases both its dataset and its split, so our
re-analysis covers one published gain.

\subsection{Recommendations for fusion studies and dataset releases}

A fusion study on a paired image--sensor dataset should report two things. The
first is its fusion gain on a farm- or session-disjoint split. Every fusion study
needs this split, because the confound that it controls for arose through a
different mechanism in each of two datasets assembled by separate groups: on the
AI Hub dataset, readings from sensors installed per farm and matched to the images
afterwards, and on CDD, per-image files that repeat one environment series across
many images. A study
therefore cannot rule the confound out by checking for one mechanism. The second is the score of a
sensor-only model, set beside the scores of a majority-class predictor and a
timestamp classifier on the same test set and metric. The timestamp classifier is needed
beside the majority-class predictor because a sensor-only accuracy far above the
majority-class accuracy is not by itself evidence of agronomic signal: session
identity alone can produce it on a split that shares sessions. Some of the fusion studies we cite publish a sensor-only score, but none sets
a majority-class score beside it, so a reader cannot tell how far those
scores exceed the majority-class score. \citet{berrocal2026} give their soil-moisture-only baseline its own subsection
and table, yet even they interpret that baseline's accuracy as ``a meaningful but
limited level of discrimination'' relative to their image model rather than to a
majority-class predictor. 

Dataset releases can let readers check for session confounding before anyone
trains a model. Two disclosures would tell a reader from the release alone how
many images share each sensor reading: the count of distinct sensor readings, and the
unit over which a reading is constant, such as the session.

\subsection{Limitations and future work}

Neither dataset that we examined has sensor readings known to carry agronomic
signal, so we have shown that the diagnostic finds confounded datasets but not
that it separates them from unconfounded ones. Showing that would need a dataset
that records a variable known to drive the disease at the plant and time of each
image. The diagnostic also cannot certify a
dataset on which it finds no confound, because a partial input that fails to
predict the label does not show that no label cue exists
\citep{feng2019misleading}. The timestamp baseline uses random forests and the
fusion experiment one fusion architecture, the CDD study's late fusion, so neither
sets an upper bound on what a stronger learner could extract from the sensor
readings. A date-disjoint test that settles whether the AI Hub fusion gain in
macro-F1 exceeds zero on held-out dates needs more random seeds than we ran. On CDD,
we ran no split that holds out whole photo dates, which would remove the photo-date
confound for every class but the \cddSingleDateClasses{} photographed on a single
date. Both
datasets come from Korean farms and greenhouses, so the extent of the confound in
datasets collected elsewhere remains to be measured.

\section{Conclusion}
\label{sec:conclusion}

Studies that fuse crop images with environmental sensor readings report gains on
paired datasets without establishing what the readings measure. We tested whether
the readings identify the session with a diagnostic that reads no image. In both Korean datasets
that we examined, nearly every image shares its sensor reading with other images.
A timestamp alone is comparable to or better than the soil-chemistry readings in
every crop of the AI Hub dataset that we compared. On CDD, neither the timestamp
nor the environment series predicts the class better than the other in macro-F1
under either split that we draw. In
the split of a published CDD fusion study, \cddLeak{}\% of the test images share
their environment series with training images. On that split, a timestamp alone matches the macro-F1 that the
study reports for its fusion model. A late-fusion gain on the AI Hub
dataset replicates in distribution when the models are trained for the CDD
study's \matchedEpochs{} epochs, but whether it exceeds zero in macro-F1 on held-out
dates remains unresolved at the seed count we ran. On both datasets, the diagnostic
finds that the readings mainly identify the session, so a fusion study can show what
its readings add to the image only on a farm- or session-disjoint split.

\section*{Data availability}
\label{sec:availability}

We do not redistribute the AI Hub dataset
because, although it carries a CC-BY-SA-4.0 label, its download channel is
restricted to approved domestic applicants and requires modified versions to be
submitted to the provider before external release. Requests to access it should be directed to AI
Hub \citep{aihub2023subtropical}. The CDD dataset and the CDD study's code are not
redistributed either.
Every value that the CDD study reports is cited to the study.

\section*{Use of large language models}

Large language models (LLMs; Claude Opus 4.8 and Claude Opus 5.5, Anthropic) were
used for the following tasks: (1) drafting manuscript prose from author-provided
outlines and structured notes, (2) editing for clarity and concision, and (3)
writing Python analysis scripts from author-specified requirements and
pseudocode. All LLM-generated content was reviewed, verified, and approved by
the author. The LLMs did not contribute to study design, data collection, or
interpretation of results.

\bibliographystyle{unsrtnat}
\bibliography{refs}

\end{document}